\pdfoutput=1

\documentclass[11pt]{article}

\newcommand{\eat}[1]{}

\usepackage{latexsym}
\usepackage{amsfonts}
\usepackage{amsmath}
\usepackage{xcolor}
\usepackage{colortbl}
\usepackage{epsfig}
\usepackage{xspace}
\usepackage{graphicx}
\usepackage{subfigure}
\usepackage{paralist}
\usepackage{enumerate}
\usepackage[color,matrix,arrow,all]{xy}
\usepackage{comment}
\usepackage{booktabs}
\usepackage{balance}
\usepackage{stmaryrd}
\usepackage{pifont}
\usepackage{hhline}
\usepackage{listings}
\usepackage{array}
\usepackage{float}
\usepackage[flushleft]{threeparttable}

\usepackage{mathrsfs}
\usepackage{makecell}
\usepackage{xparse}
\usepackage{wrapfig}

\usepackage{epsfig}
\usepackage{multirow}
\usepackage{url}

\usepackage{multirow}
\usepackage{natbib}
\usepackage{graphicx}

\usepackage{listings}
\usepackage{framed}
\usepackage{xcolor}
\usepackage{color}
\usepackage{geometry}
\usepackage{changepage}
\colorlet{shadecolor}{gray!20}

\definecolor{shadecolor}{RGB}{220,220,220}

\definecolor{inputcolor}{RGB}{255,139,35}
\definecolor{outputcolor}{RGB}{120,212,252}
\definecolor{embedcolor}{RGB}{254,127,156}
\definecolor{maskcolor}{RGB}{122,128,255}
\definecolor{ecolor}{RGB}{58,149,54}

\definecolor{highcolor}{RGB}{255,153,153}
\definecolor{midcolor}{RGB}{255,204,204}
\definecolor{lowcolor}{RGB}{204,229,255}

\usepackage{tikz}
\usetikzlibrary{shapes,decorations}
\usetikzlibrary{calc}

\usepackage[export]{adjustbox}

\definecolor{green}{RGB}{0,128,0}

\definecolor{yellow}{RGB}{255,200,18}

\newcommand{\stab}{\vspace{1.2ex}\noindent}

\newcommand{\bi}{\begin{itemize}}
\newcommand{\ei}{\end{itemize}}

\newcommand{\be}{\begin{enumerate}}
\newcommand{\ee}{\end{enumerate}}
\newcommand{\beqn}{\begin{eqnarray*}}
\newcommand{\eeqn}{\end{eqnarray*}}

\newcommand{\stitle}[1]{\stab\noindent{\bf #1}}

\newcommand{\eg}{{\em e.g.,}\xspace}

\makeatletter
    \newcommand\figcaption{\def\@captype{figure}\caption}
    \newcommand\tabcaption{\def\@captype{table}\caption}
\makeatother

\tikzstyle{mybox} = [draw=black, fill=black!5, thick,
   rectangle, rounded corners, inner sep=0pt, inner ysep=6pt]
\tikzstyle{fancytitle} =[fill=black, text=white]

\NewDocumentCommand{\nan}{ mO{} }{\textcolor{blue}{\textsuperscript{\textit{Nan}}\textsf{\textbf{\small[#1]}}}}

\NewDocumentCommand{\yang}{ mO{} }{\textcolor{green}{\textsuperscript{\textit{yang}}\textsf{\textbf{\small[#1]}}}}

\NewDocumentCommand{\zzx}{ mO{} }{\textcolor{yellow}{\textsuperscript{\textit{zzx}}\textsf{\textbf{\small[#1]}}}}

\usepackage[]{ACL}
\usepackage{amsmath}
\usepackage{graphicx}
\usepackage{times}
\usepackage{amssymb}
\usepackage{multirow}

\usepackage[T1]{fontenc}

\usepackage[utf8]{inputenc}

\usepackage{microtype}

\usepackage{inconsolata}

\usepackage[ruled,linesnumbered]{algorithm2e}
\usepackage{booktabs} 
\usepackage{siunitx} 

\newcommand{\qa}{{MEBench}\xspace}
\newcommand{\sys}{{SRAG}\xspace}
\newcommand{\reason}{{SQA}\xspace}
\usepackage{multirow}
\usepackage{mdframed}

\title{SRAG: Structured Retrieval-Augmented Generation for Multi-Entity Question Answering over Wikipedia Graph}

\author{
 \textbf{Teng Lin\textsuperscript{1}},
 \textbf{Yizhang Zhu\textsuperscript{1}},
 \textbf{Yuyu Luo\textsuperscript{1,2}},
 \textbf{Nan Tang\textsuperscript{1,2}},
\\
 \textsuperscript{1}DSA,HKUST(GZ)
 \\
 \textsuperscript{2}HKUST
\\
 \small{
    \href{mailto:email@domain}{\{tlin280, yzhu305\}@connect.hkust-gz.edu.cn}
 }
 \\
 \small{
    \href{mailto:email@domain}{\{yuyuluo, nantang\}@hkust-gz.edu.cn}
 }
}

\begin{document}
\maketitle


\begin{abstract}
Multi-entity question answering (MEQA) poses significant challenges for large language models (LLMs), which often struggle to consolidate scattered information across multiple documents. An example question might be ``What is the distribution of IEEE Fellows among various fields of study?'', which requires retrieving information from diverse sources \eg Wikipedia pages. The effectiveness of current retrieval-augmented generation (RAG) methods is limited by the LLMs' capacity to aggregate insights from numerous pages. To address this gap, this paper introduces a structured RAG (SRAG) framework that systematically organizes extracted entities into relational tables (\eg tabulating entities with schema columns like ``name'' and ``field of study'') and then apply table-based reasoning techniques. Our approach decouples retrieval and reasoning, enabling LLMs to focus on structured data analysis rather than raw text aggregation. Extensive experiments on Wikipedia-based multi-entity QA tasks demonstrate that SRAG significantly outperforms state-of-the-art long-context LLMs and RAG solutions, achieving a 29.6\% improvement in accuracy. The results underscore the efficacy of structuring unstructured data to enhance LLMs’ reasoning capabilities. 
\end{abstract}


\section{Introduction}

\begin{figure*}[t!]
\begin{center}
\includegraphics[width=1\linewidth]{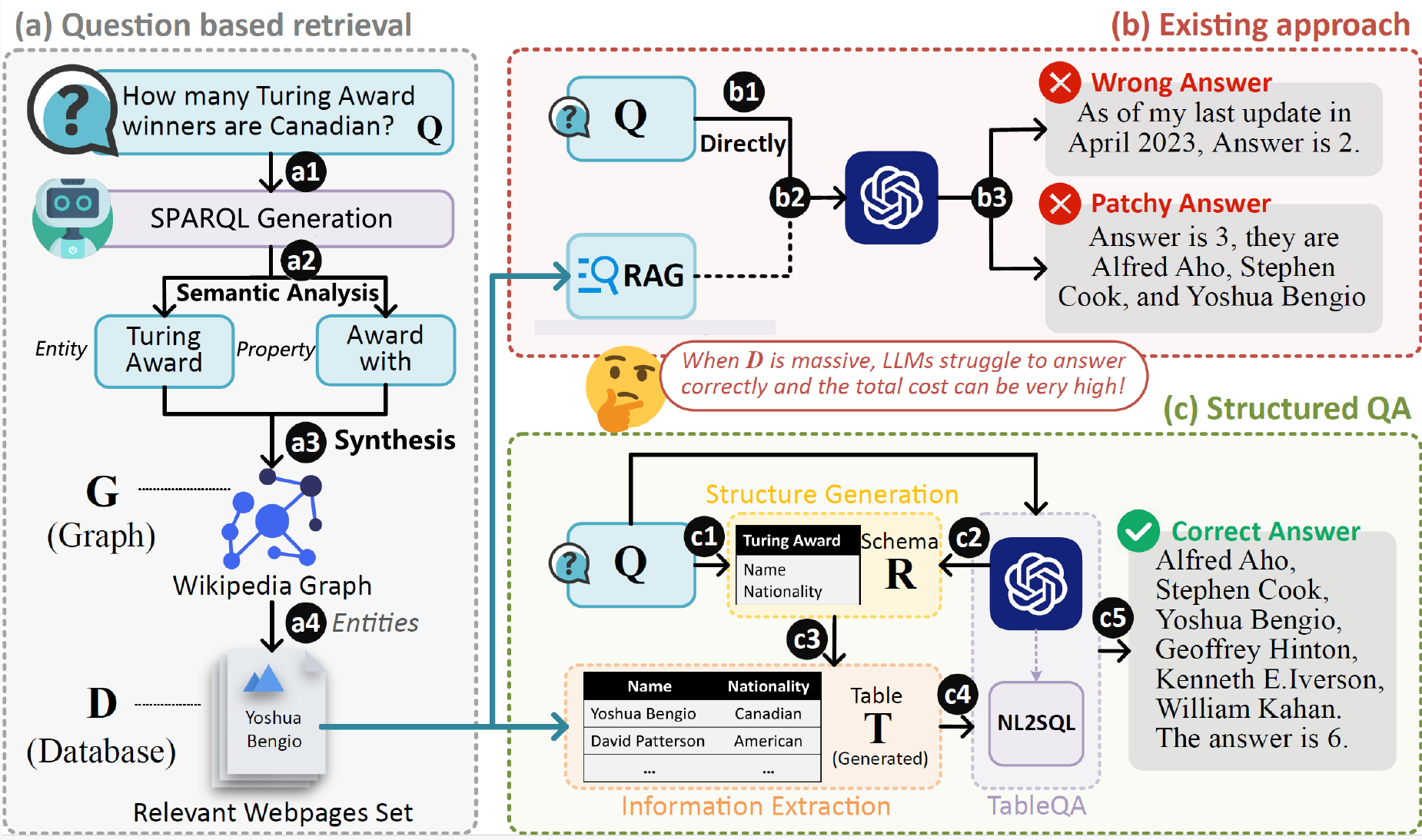}
\end{center}
\caption{An Overview of Multi-entity QA Solutions over Wikipedia Graph. (a) \textbf{Multi-Entity Retrieval}: In a1, a rough SPARQL query is generated using language model (GPT-4). In a2, we integrate the LLM's semantic parsing with Wikipedia's API and utilize verifiable query accuracy on structured Wikidata to accurately identify entities and properties. In step a3, we synthesize an exact SPARQL query. Finally, in a4, the refined SPARQL query is used to retrieve the relevant entities and web pages. (b) \textbf{Existing Reasoning Solutions}: b1 represents direct responses from LLMs, while b2 combines LLMs with RAG. (c) Our proposal: \textbf{Structured QA}. Initially, in step c1, a language model (GPT-4) is employed to analyze the question and determine the \textbf{table schema}. In c3, we utilize an \textbf{information extraction module} to populate the table. Finally, in step c4, the \textbf{TableQA} module is used to derive the final answer.}
\label{fig:example}
\end{figure*}

Recent progress in Retrieval-Augmented Generation (RAG) has enhanced how language models access external knowledge, improving applications like question answering and document integration~\citep{fan2024Asurvey}. By merging advanced retrieval methods with powerful language models, these systems have shown strong performance. However, challenges persist in accurately retrieving entity information from multi-document and heterogeneous knowledge bases. This challenge becomes especially apparent in Multi-Entity Question Answering, the challenge lies not only in recognizing and extracting relevant entities precisely from data but also in understanding the properties of these entities within the context of the query. Consider answering questions such as ``What are the capitals of countries bordering France?'' or ``How many Turing Award Winners are Canadian'' (the query $Q$ in Figure~\ref{fig:example}-a). Answering these questions requires the extraction of multiple documents, unless specific statistical analysis has been carried out by hand in advance. Existing RAG methods struggle with MEQA tasks, because useful information required to these tasks is badly scattered. This characteristic makes it difficult for existing RAG methods to accurately identify key information and perform global reasoning with noisy retrieved content.

To tackle the challenge, we propose an innovative \textbf{Structured RAG System (SRAG)} which includes two main parts: (1) \textbf{Multi-entity Semantic Retrieval} as illustrated in Figure~\ref{fig:example}-a, and (2) \textbf{Structured Question Answering (SQA)} as illustrated in Figure~\ref{fig:example}-c, to effectively address multi-entity QA. 

\paragraph{Contributions} Our notable contributions are summarized as follows.

\bi
    \item \textbf{MEBench: Specialized Benchmark for MEQA~\citep{lin2025mebench}.} We curate \qa, a standardized benchmark based on Wikipedia to evaluate the effectiveness of various approaches to address the complexities of information extraction and reasoning involving multiple entities.
    \item \textbf{Multi-entity Semantic Retrieval.} Multi-entity Semantic Retrieval innovatively enhances semantic retrieval by integrating language models with structured database. This integration enables the creation of precise SPARQL queries to effectively retrieve relevant entities and web pages from Wikipedia. Our key contribution lies in improving retrieval accuracy by leveraging contextual brilliance of language models with the rigor of structured data validation.
    \item \textbf{SQA: Module for Structured Entity Information and Reasoning.} We propose the \reason, a module for managing vast and unstructured data by extracting properties of entities and organizing information into structured tables. This module transforms textual information of entities into a format with a rigorous and accurate schema, which facilitates analysis. Our experiments demonstrate its remarkable performance, achieving SOTA results and outperforming the strongest baselines by 29.6\% in overall accuracy, while leading across all eight subtasks.
\ei

\section{Related Work}

\subsection{Retrieval-Augmented Generation}
Advanced RAG systems have evolved to incorporate pre-retrieval, retrieval, and post-retrieval strategies aimed at mitigating the limitations of basic RAG methods. In parallel, Modular RAG systems have introduced frameworks for iterative and dynamic cycles of intertwined retrieval and generation processes~\citep{gao2023retrieval}. Community summaries serve as a form of self-memory~\citep{cheng2023liftup}, enhancing generation-augmented retrieval to support future cycles of information generation. Additionally, the parallel generation of community answers from these summaries reflects an iterative~\citep{shao2023enhancing} or federated ~\citep{wang2024feb4} retrieval-generation approach. Furthermore, using a hierarchical index and summarization aligns with alternative methods, such as generating a hierarchical index of text chunks through vector clustering ~\citep{sarthi2024raptor}. These methods concentrate on semantic aggregation and overlook the structured views of multiple entities.

\subsection{Graph RAG}
Recently, several studies have explored the incorporation of graph structures into RAG systems to enhance LLMs in addressing complex question-answering tasks~\citep{edge2024local} ~\citep{panda-etal-2024-holmes}. One approach involves utilizing pre-existing knowledge graphs, from which subgraphs are extracted based on specific queries. These subgraphs are then encoded as soft prompts or converted into plain text for the generation module GraphGPT~\citep{Tang2024graphgpt} and other 
methods~\citep{he2024gretriever}~\citep{guan2023mitigating}. Another method focuses on extracting entity-relation triples from relevant text documents according to query needs, allowing for the construction of graph structures that facilitate knowledge augmentation, like GraphRAG~\citep{edge2024local} and RQ-RAG~\citep{Chan2024RQRAGLT}. These models are inadequate for addressing multi-entity questions. In response, we introduce a comprehensive Multi-Entities QA benchmark \qa based on Wikipedia Graph and provide a first-of-its-kind original system \sys.


\section{Problem Statement} 
\subsection{Wikipedia Graph}
\begin{figure*}[t!]
\begin{center}
   \includegraphics[width=0.85\linewidth]{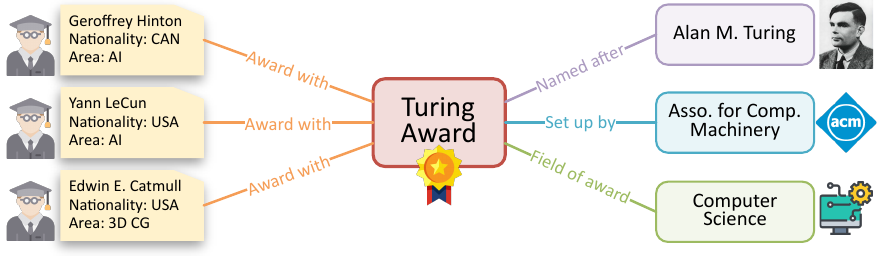}
\end{center}
  \caption{Illustration of a Wikipedia graph snippet.}
  \label{fig:prograph}
\end{figure*}

Wikipedia graph is represented as {\emph{G=(V, E, P)}}, where $V=\{v_1, v_2, ..., v_n\}$ is the set of nodes in the graph, with each node $(v_i \in V)$ representing an entity; $E$ is the set of direct edges in the graph, with each edge $(e_j (v_i, v_k) \in E)$ representing a connection (or relationships) between two nodes. An edge is a tuple $(v_a, v_b, r)$, where $(v_a, v_b \in V)$ and $r$ is the type of relationship. For example, $E=\{(v_1, v_2, r_1), (v_2, v_3, r_2), ..., (v_m, v_n, r_k)\}$.
$P$ represents the set of properties associated with both nodes and edges. 
Figure~\ref{fig:prograph} is a sample of a Wikipedia graph snippet.
\subsection{Multi-Entities Question}  

A Multi-Entities Question can be formally defined as \(Q = (t_Q, V_Q, P_Q)\), where \( t_Q \in T \) denotes the query type, with \( T = \{ t_1, t_2, \ldots, t_8 \} \) representing the set of eight predefined types. Details can be seen in Table~\ref{multiqexamples}. $V_Q$ denotes the collection of entities directly associated with the question. \( P_Q \) represents the comprehensive set of properties pertinent to the question, encompassing both node and edge properties. 


\section{System Design}
In our system, the \textbf{Multi-entity Semantic Retrieval} involves conducting a SPARQL retrieval across the Wikipedia graph to obtain relevant Wikipedia pages. Secondly, in SQA module, \textbf{table generation} begins with ``guessing'' a table schema based on the given query, followed by the extraction of information from the identified entities to populate the table. Finally, we implement an \textbf{executor} that processes the generated table to respond to the query. Next, we will elaborate on the details of each step.

\subsection{Composite SPARQL Retrieval}
Initially, we utilize GPT-4 to parse the question to construct rough SPARQL.The entities ID and properties ID contained in the rough SPARQL frequently turn out to be inaccurate. To make SPARQL valid, we deploy GPT-4 as Semantic Analysis Model to identify entities and properties. Integrating with the Wikipedia API, we get right \colorbox{black!5}{(entityID, propertyID)} pair to replacing the wrong IDs in the rough SPARQL, as illustrated in Figure~\ref{fig:example}-a2. Consequently, our system is capable of performing entity and property identification without ambiguities. 
For multi-hop queries, the Semantic Analysis process initially deconstructs the queries into sub-queries, allowing composite SPARQL retrieval step to be applied sequentially to each sub-query to identify named entities and extract properties until all sub-queries have been processed.

\subsection{Table Generation}
Although SPARQL provides aggregation functions such as ``SUM, AVG, COUNT'', etc. they are insufficient for complex statistical problems. Therefore, we build tables instead of extended graphs to support more complex algorithms and analysis.
The table generation consists of two steps: (1) Generation of schema; (2) Extracting entity information to fill the table.

\textbf{Schema Generation.}
We employ GPT-4 to systematically parsing the question to identify critical entities, attributes, and their interrelationships, which are then formalized into a structured schema. The generated schema may necessitate adjustments or refinements to align more closely with the intent of the question. For example, for the question ``How many Nobel Prizes in Physics laureates have been awarded for discoveries in Particle Physics?'', the LLM produces a schema \colorbox{black!5}{(name, YearAwarded, field)}, there are two issues, first is that \colorbox{black!5}{(field)}is oversimplified, it should be \colorbox{black!5}{(field in Physics)}. The second issue is that columns \colorbox{black!5}{YearAwarded} are redundant. Therefore, We prompt GPT-4 to critically review content to minimize oversimplification, omission of essential elements, and redundancy, and the prompt is shown in Appendix~\ref{appendix:schema}.

\textbf{Entity Information Extraction.}
In our data processing workflow, we use the LLM (Mistral-7B) to extract information from the retrieved data, populating a table where each row represents a unique entity, as shown in Figure~\ref{fig:example}-c3. This step transforms entities into structured tables for downstream table-based reasoning tasks.

\subsection{Executor}
We utilize GPT-4 to generate SQL according to the question. To increase accuracy, we include relevant information in the prompt, such as the table schema and data samples. The generated SQL are executed on the generated table to obtain results, which could be a single value or a subset of the table.



\begin{table*}[t!]
\centering
\renewcommand\arraystretch{1}
\caption{Examples of multi-entities queries.}
\vspace{.2em}
\begin{tabular}{clp{7.5cm}}
\toprule
\textbf{Categories} & \textbf{Types} & \textbf{Examples}  \\
\midrule
\multirow{2}{*}{Comparison} 
    & \multirow{1}{*}{Intercomparison}   & Which has more ACM fellow, UK or USA?\\
    \cmidrule(lr){2-3}
    & \multirow{1}{*}{Superlative} &Which city has the highest population?\\
\midrule
\multirow{8}{*}{Statistics} 
    &\multirow{1}{*}{Aggregation} &How many ACM fellow are from MIT?\\
    \cmidrule(lr){2-3} 
    &\multirow{2}{*}{Distribution Compliance} & Does the nationality of ACM fellows follow a normal distribution? \\
    \cmidrule(lr){2-3} 
    &\multirow{2}{*}{Correlation Analysis} & Is there a linear relationship between number of events and records broken in Olympic Games?\\
    \cmidrule(lr){2-3} 
    &\multirow{3}{*}{Variance Analysis} & Do the variances in the number of participating countries and total events in the Summer Olympics differ significantly?\\
\midrule
\multirow{5}{*}{Relationship} 
    & \multirow{2}{*}{Descriptive Relationship} &Is there a relationship between the year of ACM fellowship induction and the fellows' areas of expertise?\\
    \cmidrule(lr){2-3} 
    & \multirow{3}{*}{Hypothetical Scenarios} & If China wins one more gold medal, will it overtake the US in the gold medal tally at the 2024 Olympics?\\
\bottomrule
\end{tabular}
\label{multiqexamples}
\end{table*}

\begin{table*}[t!]
  \centering
    \caption{Statistics of \qa benchmark.}
    \vspace{.2em}
  \begin{tabular}{lccc}
    \toprule
    \textbf{Categories}  & \textbf{\qa -train} & \textbf{\qa -test}& \textbf{\qa -total} \\
    \midrule
    \#-Queries  &3406 &1374 &4780 \\
    \#-one-hop Q    & 1406 &606 & 2012\\
    \#-multi-hop Q    & 1322& 768& 2090\\
    Ave. \#-entities /Q   &460 &391 & 409\\
    \#-Topics  &165 &76 & 241\\
    \#-Comparison  &1107 &438 & 1545\\
    \#-Statistics &1440 &585 & 2025\\
    \#-Relationship &859 &351 & 1210\\
   \bottomrule
  \end{tabular}
  \label{querycategories}
\end{table*}

\section{Experiment} 
\subsection{Experiment Setup}
\paragraph{\qa Benchmark.}
It is a specialized benchmark designed to evaluate systems addressing multi-entity QA. The benchmark comprises 4,780 methodically structured questions partitioned into two subsets: a training set (3,406 questions) for model fine-tuning and a test set (1,374 questions) for rigorous evaluation. These questions are systematically categorized into three primary categories, further divided into eight distinct types (see Table ~\ref{multiqexamples}), ensuring broad coverage of real-world multi-entity reasoning scenarios. Table~\ref{querycategories} details comprehensive statistics of the benchmark. 
\paragraph{Baselines.}
For open-source LLMs, we conduct experiments using the representative Meta-Llama-3-8B-Instruct~\citep{Llama3_2024} and  apply QLoRA~\citep{dettmers2023qlora} to fine-tune it with the training set of \qa. For proprietary LLMs, we select the widely recognized GPT models, including GPT-3.5-turbo~\citep{ouyang2022training} and GPT-4~\citep{achiam2023gpt}. Additionally, we incorporate RAG across all vanilla baseline models for comparative analysis and evaluation of the model's capacity to integrate and leverage external data sources.
\paragraph{Evaluation Metrics.}
We adopt Accuracy ($Acc$) as the primary metric to assess the performance of LLMs on \qa tasks. For the subcategories of Variance Analysis, Correlation Analysis, and Distribution Compliance within the Statistics tasks shown in Table~\ref{fig:example}, we focus solely on prompting LLMs to identify relevant columns and applicable methods, evaluating the accuracy of their selections instead of the computational results, as LLMs' abilities in precise calculations are not the central focus of this study.

\subsection{Results and Analysis}

\begin{table*}[t!]
  \centering
  \renewcommand\arraystretch{1}
  \caption{
   Experimental results for \qa. 
  }
  \vspace{.2em}
  \begin{tabular}{lcccc}
    \toprule
    \multirow{2}{*}{\textbf{Models}} 
        & \multicolumn{4}{c} {\textbf{Accuracy}}\\
        \cmidrule(lr){2-5}
        & Comparison &Statistics &Relationship &Overall\\
    \midrule
    GPT-3.5-turbo &0.105 	&0.198 	&0.476 	&0.239 \\
    GPT-3.5-turbo + RAG & 0.605 	&0.260 	&0.476 	&0.425 \\
    GPT-4 & 0.199 	&0.289 	&0.507 	&0.316 \\
    GPT-4 + RAG & 0.763 	&0.410 	&0.687 	&0.593 \\
    Llama-3-Instruct & 0.046 	&0.118 	&0.256 	&0.130 \\
    Llama-3-Instruct + RAG & 0.447 	&0.181 	&0.410 	&0.325 \\
    FT Llama-3-Instruct & 0.046 	&0.253 	&0.259 	&0.189 \\
    FT Llama-3-Instruct + RAG & 0.687 	&0.448 	&0.573 	&0.556 \\
    \hline
    \textbf{\sys (Ours)} & \textbf{0.934} &\textbf{0.908} &\textbf{0.803}	&\textbf{0.889} \\
    \bottomrule
  \end{tabular}
  \label{Accuracy}
\end{table*}

\begin{figure*}[t!]
\begin{center}
\includegraphics[width=1\linewidth]{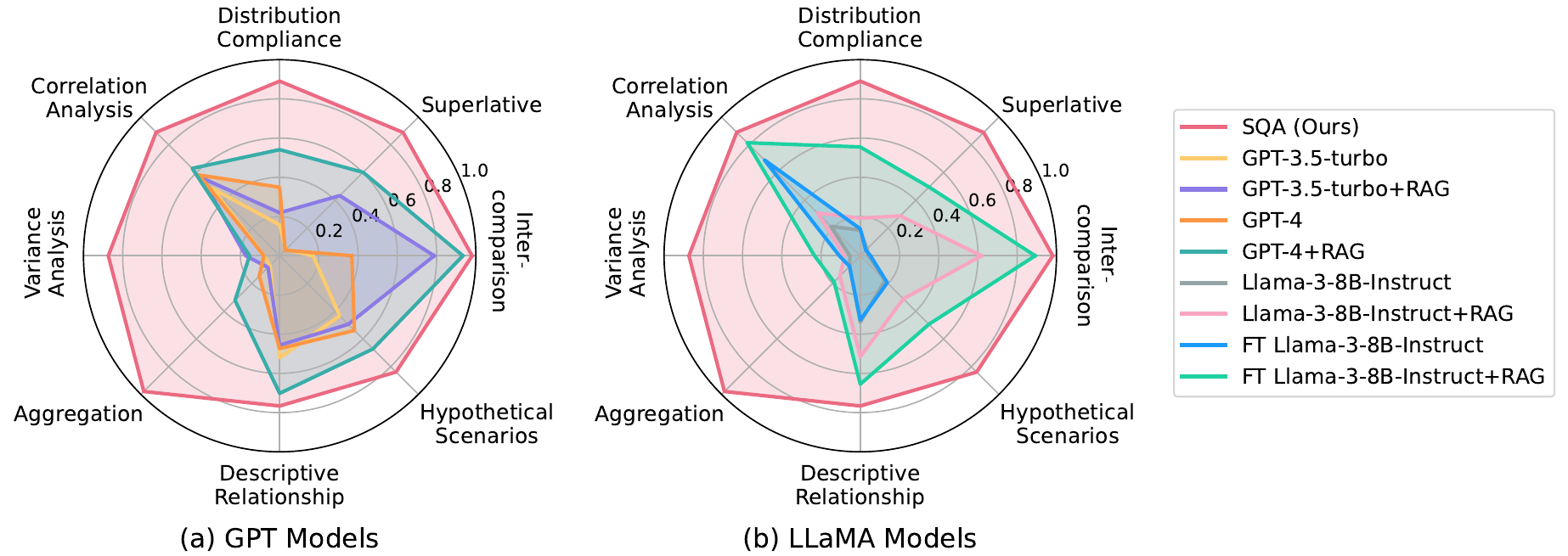}
\end{center}
  \caption{Experimental results for eight types queries of each model.}
  \label{fig:acc}
\end{figure*}

Various models exhibit notable variations in performance on \qa. 
Table~\ref{Accuracy} presents experimental results alongside overall accuracy on \qa, and Figure~\ref{fig:acc} shows accuracy on eight further-divided types.

\stitle{Performance of \sys and Baselines.}
Compared to baselines, our \sys significantly improves overall accuracy, reaching 88.9\% and increasing the best baseline (GPT-4 + RAG) by \textbf{29.6\%}. Our approach outperforms the accuracy in the relational and comparative query types by 11.6\% and 17.1\%, respectively, while achieving a remarkable improvement of 46\% for statistical query types.

\stitle{Fine-grained Performance on Sub-tasks.}
Figure~\ref{fig:acc} shows that vanilla LLMs perform well in correlation analysis and descriptive relationship sub-tasks, while RAG significantly improves intercomparison and superlative tasks. However, neither fine-tuning nor RAG overcomes challenges in variance analysis and aggregation tasks, while our proposed \sys achieves superior accuracy of 87.3\% and 97.9\%.

\stitle{Errors Analysis for \sys.} \label{sec:error_analysis}
We sample and analyze the output of the \sys system. It faces two challenges which are listed below. 

\bi
    \item \textbf{Relation semantic parsing.} 
    The semantic parsing model in SPARQL retrieval effectively recognizes entities but struggles with relationship identification, leading to challenges in graph retrieval and negatively affecting the performance of RAG-based approaches, including SQA. For example, in the query ``How many US presidents have served more than one term in office?'' The model incorrectly identifies the relationship as ``instance of'' rather than the correct ``position held'', leading to erroneous results.

    \item \textbf{Insufficient information extraction.} We also identified errors in \sys's information extraction during the table-filling phase. An analysis of more than 2,000 table filling instances reveals that these errors occur primarily as omissions (albeit with a low probability of approximately 0.1\%). A new challenge is the appearance of multi-word synonyms within the same column, such as ``US'' and ``America'', which negatively affects the accuracy of SQL execution such as ``SELECT''.
\ei


\section{Conclusion}
Our research presents a novel framework, Structured RAG system (\sys), to address the complexities involved in multi-entity question answering (QA) from Wikipedia. Existing methods, particularly those employing RAG alongside LLMs, often fall short in effectively aggregating and reasoning over information scattered across multiple Wikipedia pages. By leveraging the inherent structure of wiki-graph for multi-entity retrieval and introducing a system to organize extracted entities into a relational table format, \sys significantly enhances the performance of multi-entity questions answering. The exhaustive experiments conducted underscore the superior performance of \sys in overcoming the limitations of traditional RAG-based solutions. This research not only presents a more effective methodology for multi-entity QA but also sets the stage for future explorations into improving the accuracy and efficiency of information mining from large, unstructured knowledge bases.

\section{Limitations}
The limitations identified in the error analysis in Section~\ref{sec:error_analysis} emphasize the need for better semantic understanding of queries and the use of large language models (LLMs) to improve data extraction accuracy. Another key limitation of the system is its ability to handle ambiguous queries, which in real-world scenarios can vary significantly in complexity. This requires advanced semantic inference to accurately grasp user intent, demanding the system to comprehend both explicit content and implicit context. The challenge is to balance precision and recall, allowing diverse query handling while minimizing errors. Also, we did not compare the SQL query-based table method with simpler techniques such as keyword extraction and summarization. This was based on the assumption that SQL's advantages in structured query handling are apparent. However, this overlooks the potential research value such a comparison could provide.



\bibliography{refs/custom}

\appendix
\newpage
\section{Appendix}
\label{sec:appendix}

\subsection{Hops}
In terms of Wikipedia graph systems, the term `hop' refers to a step taken along the edges of a graph from one node to another, so we consider 'hop' as a tuple $(v_a, v_b, t)$. For no hop, there is no relationship (edge) to track. Single hop track all entities have relationship $t$ to $v_a$. Multi-hop involves traversing multiple edges (hops) to find connections between nodes that are not directly linked, like track $(v_a, v_b, t_1)$ $(v_b, v_c, t_2)$.

\subsection{Prompt}
\subsubsection{Prompt for schema} \label{appendix:schema}
\begin{mdframed}
Create a table schema that comprehensively captures information about \{......\}. Ensure the schema is detailed and structured, avoiding over-simplification, missing elements, and redundancy. This schema should be structured so each row represents a unique instance, with each column capturing a distinct aspect of property details. Ensure there is no overlap in content between columns to avoid repetition.
\end{mdframed}

\subsection{Optimization} 
Two aspects of optimization are included in \sys system to enhance the overall performance:


\stitle{Model Selection.} \label{sec:model_selection} 
Model selection is straightforward yet highly effective for optimization~\citep{liu2024declarative}. The \sys system comprises multiple tasks, necessitating the selection of the most suitable model for different tasks. For basic tasks, more affordable and faster LLMs can suffice, while utilization of the most advanced LLMs is essential in more complex tasks to ensure optimal performance. Specifically, \sys system employs powerful yet resource-intensive GPT-4 for tasks such as semantic analysis or generation of table schemas and SQL queries. In contrast, for more basic information extraction, we utilize open-source Mistral-7B, thereby achieving a balance between cost efficiency and functional performance.

\stitle{LLM Input/Output Control}. SplitWise~\citep{patel2023splitwise} shows that LLM inference time is generally proportional to the size of input and output tokens. Since GPT models decide the cost based on the input token, we try to minimize the input of large models. Meanwhile, we use the instructive prompt to reduce the size of the outputs generated by LLM without changing the quality of these outputs. The example of prompt is in Appendix~\ref{appendix:output}.

\subsubsection{Prompt for Output Control}\label{appendix:output}
\begin{mdframed}
...review your output to ensure it meets all the above criteria. Your goal is to produce a clear, accurate, and well-structured output. Just output the \{\}, no other word or symbol.
\end{mdframed}

\subsection{Tables}
Table~\ref{qgenerate} shows examples of topics and their entities' properties. 

Table~\ref{qtypeteplate} shows examples of question templates to synthesize queries.

\begin{table*}[ht]
\centering
\caption{
Example Topics and Their Entities Properties. 
}
\vspace{.2em}
\resizebox{\textwidth}{!}{
\begin{tabular}{lp{7cm}c}
\toprule
    \textbf{Topics}           & \textbf{Entities Properties}  & \textbf{\#-Entities}\\
    \midrule
    \multirow{1}{*}{ACM fellow} & nationality, field of study, affiliation & \multirow{1}{*}{1115}\\
    \hline
    \multirow{1}{*}{Cities of the World} &  population, geographic coordinates, altitude, GDP  &\multirow{1}{*}{7040}\\
    \hline
    \multirow{2}{*}{Presidents of the US} & term lengths, political parties, vice-presidents, birth states, previous occupations & \multirow{2}{*}{55}\\
    \hline
    \multirow{2}{*}{Chemical Elements} &  atomic number, atomic mass, boiling point, melting point, electron configuration & \multirow{2}{*}{166}\\
    \hline
    \multirow{2}{*}{Summer Olympic Games} & host cities, number of participating countries, total number of events, medal tally, records broken & \multirow{3}{*}{35}\\
    \hline
    \multirow{2}{*}{Nobel Prize in Chemistry} &  categories, year of award, country of origin, field of contribution.& \multirow{2}{*}{194}\\
\bottomrule
\end{tabular}
}
\label{qgenerate}
\end{table*}

\begin{table*}[ht]
\centering
\renewcommand\arraystretch{1}
\caption{Template example for queries generated by the LLM (GPT-4).}
\vspace{.2em}
\begin{tabular}{clp{7.5cm}}
\toprule
\textbf{Categories} & \textbf{Types} & \textbf{Template Examples}  \\
\midrule
\multirow{2}{*}{Comparison} 
    & \multirow{1}{*}{Intercomparison}   & Which has high [property], [entity A] or [entity B]?\\
    \cmidrule(lr){2-3}
    & \multirow{1}{*}{Superlative} & Which [entity] has the highest/lowest [property]?\\
\midrule
\multirow{6}{*}{Statistics} 
    &\multirow{1}{*}{Aggregation} & How many [entities] have [specific property value]?\\
    \cmidrule(lr){2-3} 
    &\multirow{1}{*}{Distribution Compliance} & Does [property] follow a normal distribution? \\
    \cmidrule(lr){2-3} 
    &\multirow{2}{*}{Correlation Analysis} & Is there a linear relationship between [property A] and [property B]?\\
    \cmidrule(lr){2-3} 
    &\multirow{2}{*}{Variance Analysis} & Are the variances in [property A] and [property B] significantly different?\\
\midrule
\multirow{3}{*}{Relationship} 
    & \multirow{1}{*}{Descriptive Relationship} & How is [entity A] related to [entity B]?\\
    \cmidrule(lr){2-3} 
    & \multirow{2}{*}{Hypothetical Scenarios} & What would be the impact if [entity A] collaborates with [entity B]?\\
\bottomrule
\end{tabular}
\label{qtypeteplate}
\end{table*}

\subsection{Automated QA Generation and Validation} 
We extract the introductory paragraph of textual content for each entity from Wikipedia, akin to an abstract of the entity's page, to derive relevant property values. The preprocessing of graph node properties is conducted using GPT-4. GPT-4 is deployed to generate the essential properties of key entities for each topic, and subsequently, property values are extracted from the respective web pages of these entities. This process culminates in the formation of property tables. An illustrative example of the topics and entities' properties is provided in Appendix Table~\ref{qgenerate}.

When questions or queries are posed, the \sys system efficiently navigates the graph by utilizing both the connections (edges) and the nodes along with the associated property tables to retrieve relevant information. The property tables, which contain attributes and values related to the entities within the graph, serve as a comprehensive and structured data source that can be queried alongside the graph structure. This dual approach facilitates thorough analysis, as it takes into account both the relational context (the connections among entities) and the specific properties of the entities involved.
Moreover, such automated process benefits from low labor costs due to automation and optimization within the graph database system, reducing the need for time-consuming and error-prone manual data processing and analysis.

\subsection{Quality Control of Questions} 
We devise several strategies to ensure the integrity and effectiveness of questions.

\bi
    \item \textbf{Question Templates.} The use of templates ensures that every question is crafted with a clear structure, making it easier for respondents to understand and answer them accurately. For relationship and complex statistic questions, we turn the questions in a closed-ended style, as they require a specific response of either ``yes'' or ``no'', which makes the answer in a standardized format. 
    We meticulously prepare all question templates, with examples in the Appendix Table~\ref{qtypeteplate}. 

    \item \textbf{Question Refinement.} 
    After the initial development phase, each question undergoes a refinement process utilizing GPT-3.5-turbo. This stage is essential for improving the clarity, relevance, and neutrality of the questions. It also includes a thorough review to identify and mitigate any potential bias, contributing to minimizing misunderstandings and elevating the overall quality of the questions.

    \item \textbf{Manual review.} We assess the questions for accuracy, ensuring they are factually correct and relevant to our purpose. Manual reviews can also provide insights into whether the questions are likely to effectively elicit the intended information from answers, thereby contributing to the reliability and validity of the benchmark.
\ei

\subsection{Baseline Performance.}
Introducing RAG significantly improves overall performance, particularly in comparison tasks, while fine-tuning LLaMA-3-Instruct alone does not yield substantial gains without RAG. On \qa, open-source models like LLaMA-3-Instruct, even with RAG, can't match proprietary models like GPT-4, which achieves a 59.3\% accuracy compared to LLaMA-3-Instruct's 31.6\%.


\end{document}